\documentclass[journal,article,submit,moreauthors,pdftex]{Definitions/mdpi} 
\usepackage{ulem}
\usepackage[ruled,vlined]{algorithm2e}
\usepackage{times}
\usepackage{helvet}
\usepackage{courier}
\usepackage[ruled,vlined]{algorithm2e}
\usepackage{multibib}
\usepackage{hyperref}
\usepackage{graphicx}
\usepackage{verbatim}
\usepackage{multirow}
\usepackage{MnSymbol}
\usepackage{amsfonts}
\usepackage{amsmath}
\usepackage{subcaption}
\usepackage{epstopdf}

\firstpage{1} 
\pubvolume{1}
\issuenum{1}
\articlenumber{0}
\pubyear{2021}
\copyrightyear{2020}
\datereceived{} 
\dateaccepted{} 
\datepublished{} 
\hreflink{https://doi.org/} % If needed use \linebreak
\Title{Polarity and Subjectivity Detection with Multitask Learning and BERT Embedding}

\TitleCitation{Title}

\Author{Ranjan Satapathy $^{1}$, Shweta Pardeshi $^{2}$ and Erik Cambria $^{3,}$*}

\AuthorNames{Ranjan Satapathy,Shweta Pardeshi and Erik Cambria}

\address{%
$^{1}$ \quad Graphene AI; ranjan@graphenesvc.com\\
$^{2}$ \quad Granular AI; shweta@granular.ai\\
$^{3}$ \quad Nanyang Technological University; cambria@ntu.edu.sg\\
}

\corres{Correspondence: cambria@ntu.edu.sg}

\abstract{Multitask learning often helps improve the performance of related tasks as these often have inter-dependence on each other and perform better when solved in a joint framework. In this paper, we present a deep multitask learning framework that jointly performs polarity and subjective detection. We propose an attention-based multitask model for predicting polarity and subjectivity. The input sentences are transformed into vectors using pre-trained BERT and Glove embeddings, and the results depict that BERT embedding based model works better than the Glove based model. We compare our approach with state-of-the-art models in both subjective and polarity classification single-task and multitask frameworks. The proposed approach reports baseline performances for both polarity detection and subjectivity detection.}

\keyword{Multitask Learning; Polarity Detection; Subjectivity Detection; Deep Learning} 

\begin{document}

% \linenumbers

\nolinenumbers
\section{Introduction}
Natural language processing (NLP) research intends to create artificially intelligent behaviour in text-related tasks. %However, most of the existing methods are unable to understand the context of words used. 
	In order to accurately extract and manipulate text meaning, an NLP system must have access to a notable amount of knowledge about the world and the domain of discourse. With the data, the NLP system is reliant on the extraction of meaning from text which has resulted in an exponential interest in NLP tasks like sentiment analysis~\cite{Pang05,cambria2016affective}, microtext normalization~\cite{satapathy2020review} and others. The tasks in NLP are interrelated and could benefit from sharing each other's learning. 
	
	Multitask learning (MTL)~\cite{caruana1997multitask} has displayed remarkable success in the field of image recognition. This success can be primarily attributed to learning shared representations from multiple supervisory tasks. Extending MTL's success to NLP, we propose an MTL model to extract both sentiment (i.e., positive or negative) and subjectivity (i.e., subjective or objective) of a sentence. In multitask framework, we aim to leverage the inter-dependence of these two tasks to increase the confidence of individual task in prediction, e.g., information about sentiment can help in the prediction of subjectivity and vice-versa. For sentence-level classification, the neutral class cannot be ignored because an opinion document can contain many sentences that express no opinion or sentiment. A sentence is opinionated if it expresses or implies a positive or negative sentiment. A sentence is non-opinionated if it expresses or implies a neutral sentiment.

	\begin{figure}
		\centering
		\includegraphics[width=0.7\textwidth]{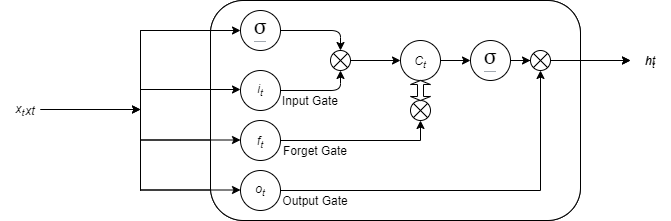}
		\caption{LSTM cell}
		 \label{fig:model-lstm2}
	\end{figure}

	Recently, BERT~\cite{devlin2019bert} has caused a stir in the NLP community by presenting state-of-the-art results in a wide variety of NLP tasks. BERT's critical technical innovation is applying the bidirectional training of Transformer, a popular attention model, to language modelling. In our proposed framework, we use BERT as an embedding to our input sentences. 
	
	We propose an MTL-based framework for polarity and subjectivity detection. Both the tasks are interrelated. The results show that our MTL framework surpasses baselines in both tasks. We evaluate our proposed approach on subjective and polarity datasets~\cite{Pang02,Pang04,Pang05}.
	%\footnote{\url{http://www.cs.cornell.edu/people/pabo/movie-review-data/}}

	\section{Related Work}
	Generic MTL~\cite{stein1956,caruana1997multitask} has a rich history in machine learning. It has widespread applications in other fields, such as
	genomics~\cite{obozinski2010}, NLP~\cite{collobert2008unified,collobert2011natural,liu2015representation,bansal2016ask} and computer vision~\cite{yim2015rotating,torralba2007sharing,misra2016cross}. 
	
	Sentiment analysis, in particular, has seen growth in multitasking usage~\cite{balikas2017multitask,majumder2019sentiment}. ~\cite{liu2017adversarial} uses MTL for adversarial text classification, mitigating the shared and private latent feature spaces from interfering with each other. ~\cite{kochkina2018all} proposes an MTL approach that allows joint training of the primary and auxiliary tasks, improving the performance of rumour verification. 
	%Subjectivity detection work --------
	
	Subjectivity detection is an NLP task
	that aims to remove ``factual'' or
	``neutral'' content (that is, objective text that does not contain any
	opinion) from online reviews. Subjectivity detection is a subtask of sentiment analysis~\cite{cambria2017sentiment}. Authors~\cite{mishra2018cognition} propose a multitask deep neural framework for document-level sentiment analysis that learns to predict the overall sentiment expressed in the given input document as the primary task.  Simultaneously learning to predict human gaze behaviour and auxiliary linguistic tasks like part-of-speech and syntactic properties of words in the document as the secondary task. Extracting subjective text segments poses a tremendous challenge that only a few works have attempted.~\cite{Pang04} applied a graph-min-cut based technique to separate the subjective portion of the text from the irrelevant objective portions. %~\cite{chaturvedi2018bayesian} introduced a novel architecture for filtering out neutral content in a time- and resource-effective manner. 
~\cite{chaturvedi2018bayesian} proposed BNELM model augments the standard recurrent neural network structure to generate a predictor that can take advantage of the beneficial properties of extreme learning machine and Bayesian networks. Emotion recognition is a task very close to sentiment classification. Authors~\cite{rashkin2019towards} used Bidirectional Encoder Representations from Transformers (BERT) to detect emotions and applied it to a dialogue system.

	\section{Task Definition}
	The MTL paradigm provides an effective platform for achieving generalization. The inspiration is that if the tasks are related, the model can learn jointly, taking into account the shared information, which is expected to increase its generalization ability. Various tasks can exploit the inter-relatedness to improve individual performance through a shared representation. Overall, it provides three principal benefits over the single-task learning paradigm:
	\begin{enumerate}
		\item it helps in achieving generalization for multiple tasks; 
		\item each task improves its performance in association with the other participating tasks; and
		\item offers reduced complexity because a single system can handle multiple problems or tasks simultaneously.
	\end{enumerate}
	MTL uses the following methods to help the model focus on the important features and ignore the noise~\cite{caruana1997multitask} :
	\begin{itemize}
		\item \textbf{Implicit data augmentation :}
		Learning just one task carries the risk of overfitting that task while learning jointly enables the model to obtain a better representation through averaging the noise patterns. MTL effectively increases the sample size we are using to train our model by sharing the learnt features. 
		\item \textbf{Attention focusing :} If a task is very noisy or data is insufficient and high-dimensional, it can be challenging for a model to distinguish between relevant and irrelevant features. 
		\item \textbf{Eavesdropping :} We can allow the model to eavesdrop through MTL, i.e., tasks challenging to learn for one model are learnt through the other model.
		\item \textbf{Representation bias :}
		MTL biases the model to prefer representations that other tasks also prefer, which helps the model to generalize to new tasks in the future. 
	\end{itemize}
	Subjectivity classification classifies sentences into two classes, subjective and objective~\cite{wiebe1999development}. An objective sentence states some factual information, whereas a subjective sentence expresses personal feelings,
	views, judgments, or beliefs. We explore this relation in our MTL-based framework. 
	
	Sentiment classification classifies sentences into two classes, positive and negative. If a sentence is classified as subjective or opinionated, we determine whether it expresses a positive or negative opinion, making sentiments and subjectivity closely related. We solve two tasks with a single network. Given a sentence sentence $s_{i}$ , we assign it both a sentiment tag (positive/negative) and a subjective tag (subjective/objective).
	
	\begin{table*}[t!]
		\centering
		\begin{tabular}{|c| c| c| c| c| c| c|} 
			\hline
			Dataset & Train & Dev & Test & Max Length & Avg. Length & Vocabulary \\ [0.5ex] 
			\hline
			POL & 7.2K & 800 &2K & 40 & 15 & 16.5k\\ [1ex] 
			\hline
			SUBJ & 7.2K & 800 &2K & 85& 17& 18.5k\\ [1ex] 
			\hline
			%21 nd 21 k
		\end{tabular}
		\caption{Statistics of the datasets used in this paper after preprocessing and output of keras tokenizer.}
		\label{table:1}
	\end{table*}

	\section{Multitask learning (MTL) based framework}
	MTL is an approach to inductive transfer that improves generalization by using the domain information contained in the training signals of related tasks as an inductive bias. We use MTL, where a single framework performs two classification tasks, i.e., subjectivity detection and polarity classification simultaneously. We have used hard parameter sharing, the most commonly used approach to MTL. It is generally applied by sharing the hidden layers between all tasks while keeping several task-specific output layers. In our experiments, we have used two different datasets and shared their information via Neural Tensor Network (NTN). The NTN helps the model share intrinsic details of each task to one another. Our model shows that the information across dataset for related tasks can be helpful to understand the task-specific features.
	
	%We assign a sentiment tag (pos/neg/) and a subjectivity tag (yes/no) to each sentence. 
\begin{comment}
    	
	\begin{figure}[h]
	  \centering
	  \includegraphics[scale = 0.3]{modelbest.png}
	  \caption{Architecture of our framework}
	  \label{fig:model}
	\end{figure}
	
\end{comment}

	\begin{figure}[ht!]
		\centering
		\includegraphics[width=0.65\textwidth]{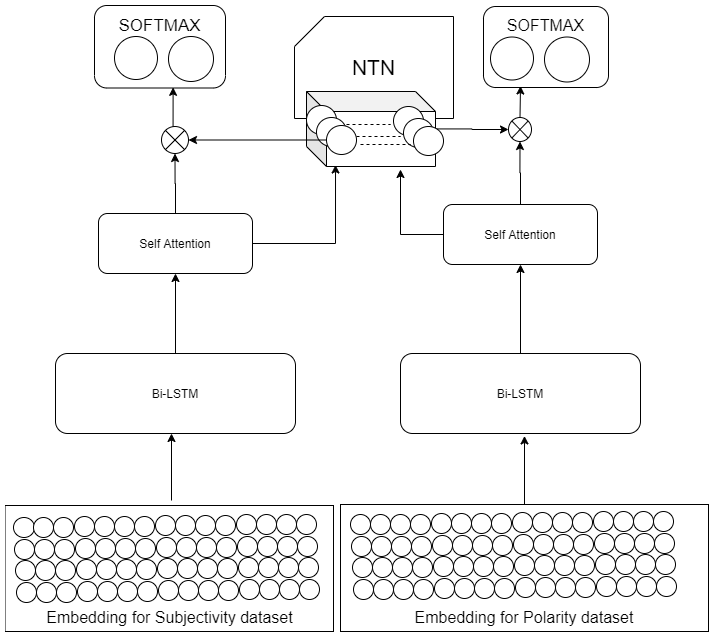}
		\caption{Framework of our proposed model}
		\label{fig:model}
	\end{figure}
	
	\begin{algorithm}
		\SetAlgoLined
		\KwResult{Class}
		
		\begin{enumerate}
			\item $E_{BERT_{T_i}}$ = BERT(S)\;
			\item $L_{T_i}$ = BILSTM($E_{BERT_{T_i}}$)\;
			\item $F_{T_i}$ = TDFC($L_{T_i}$)\; %Time distributed Fully connected
			\item $F_{T_i}$ = Drop($F_{T_i}$)\;
			\item $SA_{T_i}$ = Attention($F_{T_i}$)\; 
			\item $D_{T_i}$ = FC($SA_{T_i}$)\;
			\item $D_{T_i}$ = Drop($D_{T_i}$)\;
			\item $Fn_{T_i}$ = Flatten($D_{T_i}$)\;
			\item $X_{T_i}$ = FC($Fn_{T_i}$)\;
			\item $N$ = NTN([$Fn_{T_1} \bigoplus Fn_{T_2} $])\;
			\item $C_{T_i}$ = $X_{T_i}\bigoplus N$\;
			\item $Y_{T_i}$ = FC($C_{T_i}$)\;
			
		\end{enumerate}
		
		\KwResult{BERTEmbedding}
		initialization\;
		\begin{enumerate}
			\item Token = BERTTokenizer(S)\;
			\item id = Map(Token, ID)
			\item S-new = Pad(S, maxlen)
			\item embedding = transformer(S-new)
		\end{enumerate}
		%\KwResult{Tensor Fusion }
		% initialization\;
		% \begin{enumerate}
		%   \item step 1\;
		%  \item step 2
		% \end{enumerate}
		
		\caption{Multitask Bert based Sentiment and Subjectivity (MTBERT-SS)}
	\end{algorithm}

	\subsection{Embedding}
	We implemented two different embeddings, namely, BERT and Global Vectors (Glove) based embeddings.
	\subsubsection{BERT based embedding}
	We use pre-trained BERT~\cite{Wolf2019HuggingFacesTS} and computed sentence level BERT embeddings. We used padding to normalize the variable-length input sentences to a fixed-length. The dimension of each embedding is \textbf{L x 768}, where L is the maximum length of the input text. In our experiments, we use the BERT base model with 12 encoder layers (i.e., transformer blocks) and feed-forward networks with 768 hidden units and 12 attention heads to capture transformer-based contextual word representations from both directions. The ``Attention Mask'' in BERT is simply an array of 1s and 0s indicating the presence or absence of padding tokens. This mask tells the ``Self-Attention'' mechanism in BERT not to include these PAD tokens in its interpretation of the sentence. Our proposed model uses the output of the final layer of BERT.
	
	% We have used pre-trained bert (from huggingface transformer)
	
	% Bert model consists: The embedding layer.
	% The first of the twelve transformers.
	% The output layer.
	
	%Split the sentence into tokens using bert tokenizer
	%Add special tokens to the start and end of each sentence.[SEP] sentence [CLS]
	%Map the tokens to their IDs.
	%Pad & truncate all sentences to a maximum length.
	%Explicitly differentiate real tokens from padding tokens with the "attention mask". The "Attention Mask" is simply an array of 1s and 0s indicating which tokens are padding and which aren't. This mask tells the "Self-Attention" mechanism in BERT not to incorporate these PAD tokens into its interpretation of the sentence.
	%the output of the final transformer of BERT is used by the classifier. (Usually only the first embedding corresponding to the [CLS] token is taken but we have taken embeddings of entire sentence hence their dimension is (maxlen, 768)) 
	
	%Training 
	% Divide up our training set to use 90% for training and 10% for validation.
	
	\subsubsection{Glove based embedding}
	GloVe~\cite{pennington2014glove} is a word vector technique that leverages both global and local statistics of a corpus in order to come up with a principled loss function which uses both these. We fine-tuned GLOVE with the dataset and created an embedding of \textbf{L X 300} dimension where L is the maximum length of the input text.
	
	In our experiments, we have applied both the embeddings' results.
	\subsection{Bidirectional LSTM Layer}
	
	To get the input's context-rich representation, we fed the embeddings to the Bidirectional Long Short-Term Memory (biLSTM) Layer of size $128$. The output of the LSTM layer is used for both subjectivity and sentiment analysis. 
	
	In the next layer, we fed the LSTM output to get two matrices using two fully connected layers, which are input to two different tasks.

	%%##Add equations
	\[ i_t = \sigma (W_{x^i}x_t) + W_{hi}h_{t-1} + W_{ci}c_{t-1} + B_i\]
	\[ f_t = \sigma (W_{xf}x_t + W_{hf}h_{t-1} + W_{cf}c_{t-1} + b_f)\]
	\[c_t = f_{t}c_{t-1} + i_{t} tanh (W_{xc}x_t + W_{hc}h_{t-1} + b_c)\]
	\[ o_t = \sigma (W_{xo}x_t + W_{ho}h_{t-1} + W_{co}c_{t} + b_o)\]
	\[h_t = o_{t} tanh(c_t)\]

	\subsection{Self Attention Network}
	Self-attention is an attention mechanism~\cite{vaswani2017attention} associating the various positions of a single sequence to compute its representation. We use the self-attention mechanism in the next layer as it prioritizes the words necessary for the classification.

	\begin{flalign}
	P&=\tanh(H_* W^{ATT}),\\
	\alpha&=softmax(P^T W^\alpha), \label{eqn:attn_alpha}\\
	s_*&=\alpha H_*^T, \label{eqn:sent_rep}
	\end{flalign}
	where $W^{ATT}\in \mathbb{R}^{D_t\times 1}$, $W^\alpha\in \mathbb{R}^{L\times L}$, $P\in \mathbb{R}^{L\times 1}$, and 
	$s_*\in \mathbb{R}^{D_t}$.
	In equation \ref{eqn:attn_alpha}, $\alpha \in [0,1]^L$ gives the relevance
	of words for the task, multiplied 
	by equation \ref{eqn:sent_rep} by the context-aware word representations in $H_*$.

	The output of the attention layer is flattened before feeding to the next layer, which is the Neural Tensor Network (NTN).
	
	\subsection{Neural Tensor Network (NTN)}
	
	We use Neural Tensor Network~\cite{socher2013reasoning} model to combine both the tasks. The NTN consists of a bilinear tensor layer that links the two entity vectors. %The following function computes a score of the probability of the existence of a particular relation between two entities:

	\begin{flalign*}
	\!\!
	s_{NTN}=\tanh(s_{subj} T^{[1:D_{ntn}]} s_{pol}^T+(s_{subj}\oplus s_{pol}) W+b),
	\end{flalign*}
	where,\\
	$T\in \mathbb{R}^{D_{ntn}\times D_t\times D_t}$, $W\in$ $\mathbb{R}^{2D_t\times	D_{ntn}}$, $b, s_+ \in \mathbb{R}^{D_{ntn}}$, and
	$\oplus$ stands for concatenation. The vector $s_+$ contains shared information of both sentiment and subjectivity.

	\begin{table*}[ht!]
		\centering
		%\resizebox{\columnwidth}{!}{%
		\begin{tabular}{|l|l|l|l|} 
			\hline
			&Framework & Subjective & Polarity \\ [0.5ex] 
			\hline
			\multirow{4}{*}{Baselines}&SenticNet 6~\cite{camnt6} &-& 92.8\% \\ [1ex]

			&Subjectivity detector~\cite{Pang04} &92\%&-\\
			&AdaSent~\cite{zhao2015self} & \textbf{95.5\%}& 83.1\%\\
			&CNN+MCFA~\cite{amplayo2018translations} &\textbf{95.2\%}& 83.2\%\\
			&Multitask uniform layer~\cite{liu2016recurrent} & 93.4\% & 87.1\% \\
			
			%Multitask coupled-layer& 86.5\% & 92.5\% \\
			%\hline
			&Multitask shared-layer~\cite{liu2016recurrent} & 94.1\%& 87.9\%\\
			
			\hline
			\multirow{3}{*}{BERT Embedding}&$BILSTM_{pol}$& - &77.5\%\\ [1ex]

			&$BILSTM_{subj}$& 93.5\% &-\\ [1ex]

			&$MTL_{shared NTN}$ &\textbf{95.1\%}& \textbf{94.6\%} \\ [1ex] 
			\hline
			\multirow{3}{*}{GLOVE Embedding}&$BILSTM_{subj}$& 90.7\% &-\\ [1ex] 
			
			&$BILSTM_{pol}$& - &75.9\%\\ [1ex] 
			
			&$MTL_{shared NTN}$ & \textbf{92.3\%}& \textbf{92.1\% }\\ [1ex] 
			\hline
		\end{tabular}
		%}
		\caption{Accuracy comparision of different models}
		\label{tab:results}
	\end{table*}
	\subsection{Classification}
	For the two tasks, we use two different softmax layers (task specific) for classifications detailed below.
	
	\subsubsection{Sentiment Classification}
	
	We use the output of NTN and concatenate it with the output of self-attention to determine the sentiment of the input text using the softmax layer.

	\begin{flalign*}
	\mathcal{P}_{pol}&=\text{softmax}(s_{pol}~W_{pol}^{softmax}+b_{pol}^{softmax}),\\
	\hat{y}_{pol}&=\underset{j}{\text{argmax}}(\mathcal{P}_{pol}[j]),
	\end{flalign*}
	where $W_{pol}^{softmax}\in \mathbb{R}^{D_t\times C}$, $b_{pol}^{softmax}\in \mathbb{R}^C$, 
	$\mathcal{P}_{pol}\in \mathbb{R}^C, j$ is the class value
	(0 for negative and 1 for positive), and $\hat{y}_{sen}$ is the estimated class value.

	\subsubsection{Subjectivity Classification}
	
	We use only the output of the attention layer as sentence representation for subjectivity classification since subjectivity detection is a subtask of sentiment analysis. We use the softmax layer to get the final output.

	\begin{flalign*}
	\!\!\mathcal{P}_{subj}&=\text{softmax}((s_{subj}\oplus s_+)~W_{subj}^{softmax}+b_{subj}^{softmax}),\\
	\!\!\hat{y}_{subj}&=\underset{j}{\text{argmax}}(\mathcal{P}_{subj}[j]),
	\end{flalign*}
	where $W_{subj}^{softmax}\in \mathbb{R}^{(D_t+D_{ntn})\times C}$, $b_{subj}^{softmax}\in
	\mathbb{R}^C$, $\mathcal{P}_{subj}\in \mathbb{R}^C$, $j$ is the class value
	(0 for objective and 1 for subjective), and $\hat{y}_{subj}$ is the estimated class value.

	%\begin{comment}
	
	%\end{comment}

	\begin{figure*}[ht!]
		\begin{subfigure}{.5\textwidth}
			\centering
			% include first image
			\includegraphics[width=\linewidth]{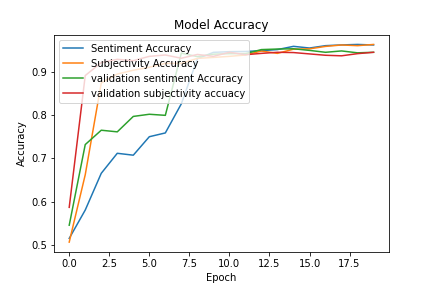}
			\caption{Accuracy graph}
			\label{fig:bert-acc}
		\end{subfigure}
		\begin{subfigure}{.55\textwidth}
			\centering
			% include second image
			\includegraphics[width=\linewidth]{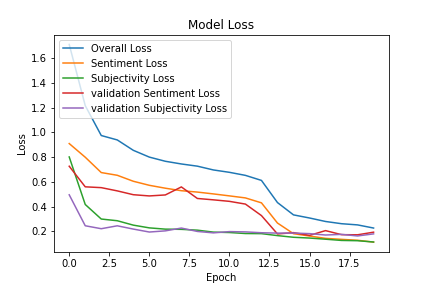}
			\caption{Loss graph}
			\label{fig:glove-acc}
		\end{subfigure}
		\caption{MTL based on BERT embedding}
		\label{fig:model-acc}
	\end{figure*}

	\section{Experiments}
	\subsection{Dataset}
	We have used the same number of sentences for both the models. Both the datasets~\cite{Pang02,Pang04,Pang05} are balanced with an equal number of classes as well.
	\begin{enumerate}
		\item POL : The dataset contains 5331 positive and 5331 negative processed sentences. We selected 5000 sentences from each class randomly i.e., 5000 positive and 5000 negative sentences.
		\item SUBJ : The dataset contains 5000 subjective and 5000 objective processed sentences. 
	
	\end{enumerate}
	Both the datasets can be downloaded from here\footnote{\url{https://www.cs.cornell.edu/people/pabo/movie-review-data/}}.
	\subsection{Baselines and Model Variants} 
	We implemented 3 different models and compare our results with 6 state of the art models as described in Table~\ref{tab:results}.
	\subsection{Hyperparameters and Training}
	\begin{enumerate}
	    \item Trainable params for the MTL model: 14,942,052
	    \item Trainable params for the individual models: 1,923,746
	\end{enumerate}
Adam~\cite{adam}  is an optimization algorithm that can be applied instead of the classical stochastic gradient descent algorithm to update network weights based on training data iteratively. The code is run for 20 epochs with 64 GB RAM and 32 GB Nvidia v100 Tesla.
	\begin{flalign*}
	\theta=&\{U^{[z,r,h]},W^{[z,r,h]},W_*,b_*,W^{ATT},\\
	&W^\alpha,T,W,b,W^{softmax}_*,b^{softmax}_*\}.
	\end{flalign*}
	
	We use categorical crossentropy ($J_*$; * is $subj$ or $pol$) as loss function:
	\begin{flalign*}
	J_*=-\dfrac{1}{N}\sum_{i=1}^N\sum_{j=0}^{C-1}{y^*_{ij}~\log \mathcal{P}_{*i}[j]},
	\end{flalign*}
	where $N$ is the number of samples which is 10K, $i$ is the index of a sample, $j$ is the class value, and
	\begin{flalign*}
	y^*_{ij}=
	\begin{cases}
	1, \text{ if expected class value of sample }i\text{ is }j,\\
	0, \text{ otherwise}.
	\end{cases}
	\end{flalign*}
	
	\section{Results and Discussions}
We divided the dataset into train, validation and test set for our experiments. The dataset was divided into train and test as 80:20 with random shuffling. The training dataset was further divided into train and validation as 90:10. We used ADAM algorithm as an optimizer with categorical cross-entropy to calculate the loss. 
	
	We implemented single task frameworks and multitask frameworks to compare the performance. The models are trained on two different datasets, so the output shows if the tasks are related, they can share essential knowledge which can benefit both the task-specific output.
	We observed that, with the addition of NTN layer, which fuses the representations of polarity and subjectivity, our MTL model performed better. In the polarity detection case, the NTN improved the accuracy by 15\% with single task framework and 10\% with MTL framework whereas in subjectivity detection case, the NTN improved the accuracy by 2-3\% with a single task framework and no improvements were found for MTL framework. 
	The comparisions with baseline also suggests an improvement in performance across both the tasks. Table \ref{tab:results} depicts the comparision of our results to baselines, a single task and multitask frameworks. The loss and accuracy graphs of our proposed BERT embedding based MTL network is shown in Figure \ref{fig:model-acc}.

	\section{Conclusion}
	Multitask learning often aids to improve the performance of similar tasks. Related tasks often have interdependence on each other and function better when solved in a joint framework. In this paper, we present a BERT based MTL framework that combines sentiment and subjective detection. In the current paper, we proposed a multilayer multitask LSTM for the main task of polarity detection and subjectivity detection. We used Neural Tensor Network to combine the task to improve the functionality of the individual tasks.
	The polarity task in MTL has improved by at least 15\% when compared to single-task performance. In comparison, the subjectivity detection task has improved only by 2-4\% in a similar setting. 
	As the tasks are related, so are the features which the model learnt to classify. However, the key takeaway from our experiment is that linguistic tasks like polarity and subjectivity detection are related tasks and, when trained on different datasets, still surpass the baselines.

% \section*{References}

\bibliography{multitask-pol_subj}
\end{paracol}
\end{document}